\documentclass[runningheads]{llncs}
\usepackage{graphicx}
\usepackage{comment}
\usepackage{amsmath,amssymb} 
\usepackage{color}
\usepackage{algorithm}
\usepackage{algorithmicx}
\usepackage{algpseudocode}

\usepackage{bm}
\usepackage{booktabs}
\usepackage[table,xcdraw]{xcolor}
\usepackage{bbm}
\usepackage{multirow}
\usepackage{hyperref}
\hypersetup{colorlinks=true,urlcolor=blue}

\usepackage{pifont}

\usepackage{rotating} 
\begin{document}
\pagestyle{headings}
\mainmatter
\def\ECCVSubNumber{2035}  

\title{Length-Controllable Image Captioning}

\titlerunning{Length-Controllable Image Captioning}
\author{Chaorui Deng\inst{1}$^*$
\and
Ning Ding\inst{1}$^*$
\and
Mingkui Tan\inst{1}$^\dagger$
\and
Qi Wu\inst{2}
}
\authorrunning{C. Deng et al.}
\institute{School of Software Engineering, South China University of Technology, China 
\email{\{secrdyz,seningding\}@mail.scut.edu.cn, mingkuitan@scut.edu.cn}
\and
Australian Centre for Robotic Vision, University of Adelaide, Australia
\email{qi.wu01@adelaide.edu.au}}
\maketitle

\begin{abstract}
The last decade has witnessed remarkable progress in the image captioning task; however, most existing methods cannot control their captions, \emph{e.g.}, choosing to describe the image either roughly or in detail. In this paper, we propose to use a simple length level embedding to endow them with this ability. Moreover, due to their autoregressive nature, the computational complexity of existing models increases linearly as the length of the generated captions grows. Thus, we further devise a non-autoregressive image captioning approach that can generate captions in a length-irrelevant complexity. We verify the merit of the proposed length level embedding on three models: two state-of-the-art (SOTA) autoregressive models with different types of decoder, as well as our proposed non-autoregressive model, to show its generalization ability. In the experiments, our length-controllable image captioning models not only achieve SOTA performance on the challenging MS COCO dataset but also generate length-controllable and diverse image captions. Specifically, our non-autoregressive model outperforms the autoregressive baselines in terms of controllability and diversity, and also significantly improves the decoding efficiency for long captions. Our code and models are released at \textcolor{magenta}{\texttt{https://github.com/bearcatt/LaBERT}}.
\let\thefootnote\relax\footnotetext{$^*$equal contribution\\$^\dagger$corresponding author}
\end{abstract}

\section{Introduction}\label{sec:intro}

Image captioning is one of the fundamental problems of computer vision which aims to generate natural language captions for images automatically. It requires not only to recognize salient objects in an image and understand their interactions, but also to describe them using natural language, which is very challenging. Most image captioning methods adopt an Encoder-Decoder framework~\cite{sutskever2014sequence,vinyals2015show,wu2017image}, where the encoder, \emph{e.g.}, a Convolutional Neural Network (CNN), first extracts features from the input image. An image caption is then decoded from the image features, one token at each time, typically using a Recurrent Neural Network (RNN).
Following this, many works \cite{huang2019aoa,li2019entangled,zhou2019unified} achieve the state-of-the-art (SOTA) performance on the challenging MS COCO dataset~\cite{chen2015microsoft}, and even outperform human performance on some evaluation metrics.

\begin{figure}[t]
\centering
\includegraphics[width=1.0\textwidth]{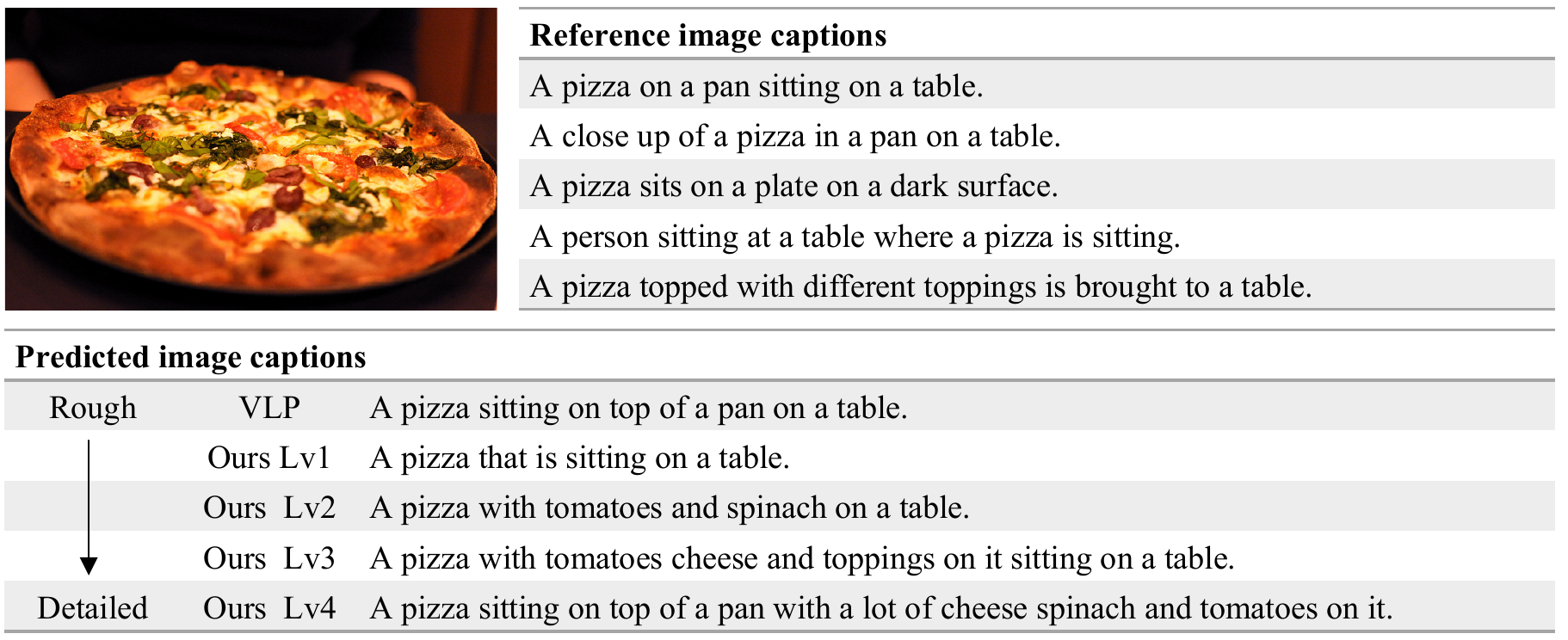}
\caption{Illustration of image captions with different lengths. 
The top left image is from the MS COCO dataset. To the right of the image are five human-annotated captions.
At the bottom, we show the image captions generated by an original VLP~\cite{zhou2019unified} model and our length-aware version of VLP.
The longest caption (Ours-Lv4) involves more instances in the image and uses more adjective phrases; while in the shortest caption (Ours-Lv1), only the salient instances are mentioned in a terse style.}
\label{fig:example}
\end{figure}

Despite their remarkable performance, many advanced image captioning approaches lack the ability to control its predictions, \emph{i.e.}, they cannot change the way they describe an image. 
See the example in Fig.~\ref{fig:example}, given an input image, although the caption generated by VLP~\cite{zhou2019unified} (a current SOTA) correctly describes the image, it also omits some informative visual concepts, such as ``cheese'' and ``tomatoes''.
If we want a more detailed description, this result would be unsatisfactory. 
Therefore, it is desired for the image captioning task if a model can be controlled to describe the image either roughly or in detail.
In this paper, we show that such an ability can be effectively acquired by directly controlling the length of the image captions.

Length is an important property of natural language since it reflects the amount of information carried by a sentence. 
As shown in Fig.~\ref{fig:example}, a longer image caption can describe an image more accurately, as it generally has higher fidelity for the image information. On the other hand, a shorter image caption can be generated and delivered more efficiently in practice, but with some extent of information loss. Nevertheless, the length property has not been explicitly exploited by previous image captioning approaches. Our work fills this gap by proposing a length-controllable image captioning model.
Specifically, we introduce a concept of ``\textbf{length level}'' which refers to a specific length range of the image captions. During training, a length level embedding is learned for each level with the training data inside the length range. During inference, the image caption generation is separated into different levels, where each level is responsible for generating image captions within the corresponding length range, conditioned on the learned length level embedding. 

Technically, this length control signal is able to be used in many existing image captioning approaches. However, they may encounter a problem when generating long image captions because of their \emph{autoregressive} nature (\emph{i.e.}, generate one token at a time conditioned on all preceding tokens): the computational complexity of autoregressive methods increase linearly as the length $L$ of the predicted caption grows, \emph{i.e}, a ${\rm{\Theta}}(L)$ complexity. To tackle this, we further develop a \textbf{non-autoregressive} model for length-controllable image captions based on the idea of iterative refinement~\cite{ghazvininejad2019mask}. 
Our non-autoregressive model is inherited from BERT~\cite{devlin2019bert}, with the input embedding layer modified to incorporate the image feature representations and the length level information. We name it as length-aware BERT (LaBERT).
Moreover, we devise a non-autoregressive decoding algorithm for LaBERT to make it be able to decode image captions within a fixed number of refine steps regardless of $L$, \emph{i.e.}, a \emph{length-irrelevant} complexity.


In our experiments, we first evaluate the effectiveness of the proposed length-controllable image captioning scheme by applying it to two most recently SOTA autoregressive image captioning methods, \emph{i.e.}, AoANet~\cite{huang2019aoa}, which employs an LSTM~\cite{Greff2015LSTM} as the decoder, and VLP~\cite{zhou2019unified}, which adopts a Transformer-based decoder. After incorporating the length information, these two methods successfully generate high-quality and length-controllable image captions. Specifically, on the challenging MS COCO dataset, our length-aware models not only outperform their original version in terms of CIDEr-D~\cite{vedantam2015cider} score on a normal length level (10 to 14 tokens), but also achieve remarkable SPICE~\cite{anderson2016spice} scores (23.0) on longer length levels (at least 14 tokens). Afterwards, we evaluate the length level control signal on our proposed non-autoregressive image captioning model, LaBERT, which achieves competitive or even better performance compared with the two autoregressive baselines on all length levels, while also significantly improves the decoding efficiency for \emph{long} captions. More importantly, we find that the proposed LaBERT has a better control precision than the two autoregressive models, and is also able to generate much more diverse image captions.

In summary, the main contributions of this paper are threefold:
\begin{enumerate}
    \item We firstly introduce the design of ``length level'' as a control signal to learn length-aware image captioning models, which can be easily integrated into existing image captioning approaches to make them capable of generating high-quality and length-controllable image captions.
    \item We devise a non-autoregressive decoder for length-controllable image captioning, which makes the decoding of long captions more efficiency. Moreover, it achieves higher control precision and produces more diverse results than the autoregressive baselines.
    \item We perform extensive experiments on different kinds of image captioning models, whether they are autoregressive or non-autoregressive, and whether they use LSTM-based decoder or Transformer-based decoder, to show the effectiveness of our proposed method.
\end{enumerate}

\section{Background and Related Works}\label{sec:pre}

\subsection{Autoregressive Image Captioning (AIC)}

Given an image $\bm{I}$, an image captioning model aims to describe $\bm{I}$ by a textual sentence $\bm{S} = \{s_i\}_{i=1}^L$, where $s_{i}$ is a token in $\bm{S}$ and $L$ is the length of $\bm{S}$. Most existing image captioning approaches operate in an autoregressive style, which factors the distribution of $\bm{S}$ into a chain of conditional probabilities with a left-to-right causal structure:
$p(\bm{S}|\bm{I}) = \prod^L_{i=1} p(s_i|s_{j<i}, \bm{I})$. 
As a result, $\bm{S}$ must be generated sequentially, \emph{i.e.}, $s_i$ cannot be generated until all preceding tokens $s_{j<i}$ are available. Assume the target image caption to be $\bm{S}^* = \{s^*_i\}_{i=1}^{L^*}$. The training of AIC models typically follows the ``Teacher Forcing''~\cite{bengio2015scheduled} scheme, which aims to maximize the likelihood of the ground-truth token $s^*_i$ given all preceding \emph{ground-truth} tokens $s^*_{j<i}$ through back-propagation:
\begin{equation} \label{eq:aic}
    \min \sum^{L^*}_{i=1} -\log p(s^*_i|s^*_{j<i}, \bm{I}).
\end{equation}
During inference, AIC models take a special \texttt{[BOS]} token as input to predict the first token $s_1$, then $s_1$ is fed into the model to obtain the next token $s_2$. Continuing like this until the special \texttt{[EOS]} token is predicted.

There have been many successful extensions to AIC approaches over the years. In~\cite{xu2015show}, the authors proposed to integrate soft and hard attention mechanisms into the decoder, which facilitates the model to learn to focus on some specific image regions when generating the token at each decoding step. Later on, \cite{rennie2017self} developed a self-critical sequence training (SCST) strategy which directly optimizes the CIDEr~\cite{vedantam2015cider} score of the predicted image captions through policy gradient~\cite{sutton2000policy} to amend the ``exposure bias'' problem in sequence modeling. Furthermore, instead of measuring attention over a pre-defined uniform grid of image regions as in~\cite{xu2015show}, Anderson \textit{et al.} \cite{anderson2018bottom} devised a bottom-up mechanism to enable the measurement of attention at object-level, which achieved the best results at that time and outperformed the second-best result by a large margin. Apart from this, some other works tried to improve the image caption quality by leveraging additional information, such as semantic attributes~\cite{yao2017boosting} and visual relations~\cite{yang2019auto,yao2018exploring,yao2019hierarchy}.
More recently, after witnessing the effectiveness of Deep Transformers~\cite{vaswani2017attention} in capturing long-range dependencies in sequence modeling, many Transformer-based AIC models~\cite{huang2019aoa,li2019entangled,zhou2019unified} have been developed to further advance the image captioning performance.

\subsection{Diverse and Controllable Image captioning}

Despite the remarkable performance achieved by current SOTA AIC models, fewer efforts have been made towards improving the diversity of image captions. In~\cite{deshpande2019fast}, the authors trained a Part-of-Speech predictor and performed sampling base on its predictions to obtain diverse image captions. Chen et al.~\cite{chen2020say} proposed to control the image captions though the Abstract Scene Graph of the image. Cornia et al.~\cite{cornia2019show} used different image regions to generated region-specific image captions. However, they rely on additional tools or annotations to provide supervision. Besides, some GAN~\cite{goodfellow2014generative}-based methods have also appeared~\cite{dai2017towards,shetty2017speaking,wang2017diverse}, most of which improve on diversity, but suffer on accuracy and do not provide controllability over the decoding process. Some other works attempted to generate image captions with controllable styles. These methods require additional training data, such as an image caption dataset with additional style annotations~\cite{chen2018factual,mathews2016senticap,shuster2019engaging}, which is scarce and expensive; or a large corpus of styled text without aligned images~\cite{gan2017stylenet,mathews2018semstyle}, which often leads to unsatisfied caption quality. 

As discussed in Section~\ref{sec:intro}, length is an important property for image captions. It is easy to acquire and is strongly associated with the structure of the image caption. Several approaches in the Natural Language Processing field have visited the length-controllable text generation setting. Kikuchi et al.~\cite{kikuchi2016controlling} proposed to control the length of output in Neural Sentence Summarization task by: 1) performing beam search without \texttt{[EOS]} token until the desired length is reached; 2) setting a length range and manually discarding out-of-range sequences; 3) feeding an embedding into the decoder in each step to indicate the \textit{remained length}; 4) incorporating the desired length information by multiplying the length with the initial hidden state. However, the first strategy may not produce completed sentences, and the results will have similar style since the model is not aware of the desired length during decoding. The second strategy may require a large beam size to obtain a valid result, and the diversity of the results is also limited. The last two strategies seeks to control the exact length of the output sentence, which is hard in practice and restricts the flexibility of the results. Moreover, the third strategy is only applicable in autoregressive text generation models. Similarly, Liu et al.~\cite{liu2018controlling} proposed to control the exact length of the output in ConvSeq2Seq models~\cite{gehring2017convolutional}. They adopted a similar way as the last strategy of~\cite{kikuchi2016controlling} that incorporate the desired length information when initializing the decoder state. As a result, they face the same problems as in~\cite{kikuchi2016controlling}.

\subsection{Non-autoregressive Text Generation}\label{sec:nat}

A common problem for autoregressive sequence generation models is that the decoding steps must be run sequentially, which prevents architectures like the Transformer from fully realizing their train-time performance advantage during inference. To tackle this, recent works in Neural Machine Translation have appealed to Non-Autoregressive Machine Translation (NAT), which attempts to make non-autoregressive predictions for the entire sequence with one forward pass of the decoder.
However, as discussed in~\cite{gu2017non}, NAT models can fail to capture the dependencies between output tokens due to the multi-modality problem, \emph{i.e.}, multiple translations are possible for a single input sentence. To deal with this, some NAT methods relaxed the one-pass restriction and adopt multiple decoding passes to iteratively refine the generated sentences~\cite{ghazvininejad2019mask,lee2018deterministic,gu2019levenshtein,stern2019insertion,wang2018semi}. To determine the length of the output, non-autoregressive approaches either predict the length of the output sentence through a length predictor, or adopt insertion/deletion modules to automatically change the length of the output.


\section{Method}

In this section, we introduce our length level embedding for length-controllable image captioning. Firstly, in Section~\ref{sec:model}, we elaborate on how the length level embedding is integrated into existing autoregressive image captioning models to endow them with the ability of length controlling. Then, in Section~\ref{sec:naic}, we introduce a non-autoregressive image captioning model that can decode image captions within a specific length range in a length-irrelevant complexity.

\subsection{Acquisition of Length Information}\label{sec:model}

Given an input image caption $\bm{S} = \{s_i\}_{i=1}^L$, to model its length information, we assign $\bm{S}$ into a specific ``length level'' with the length range \texttt{[$L_{low}$,$L_{high}$]} according to its length $L$. 
Then, we use a length level embedding matrix $\bm{W}_l \in \mathbb{R}^{k \times d}$ ($k$ is the number of levels and $d$ is the embedding dimension) to differentiate image captions on different length levels. 
Let $l$ be the length level for $\bm{S}$, and let $\bm{t}_l$ be the one-hot representation of $l$. Then, the length level embedding for tokens in $\bm{S}$ is calculated by $\bm{e}_l = \bm{W}_l^T\bm{t}_l \in \mathbb{R}^{d}$. The final representation of a token $s_i$ is constructed by adding the length level embedding $\bm{e}_l$ with its word embedding $\bm{e}_{w, s_i} \in \mathbb{R}^{d}$ and, optionally (for Transformer-based decoder), its positional embedding $\bm{e}_{p, i} \in \mathbb{R}^{d}$:
\begin{equation} \label{eq:tembed}
\bm{x}_{s_i} = \bm{e}_l + \bm{e}_{w, s_i} + \bm{e}_{p, i}.
\end{equation}
With the length level embedding $\bm{e}_l$, the length information of $\bm{S}$ is explicitly incorporated into $\bm{x}_{s_i}$. Now, given an image caption model $\mathcal{M}$, we can obtain its length-aware counterpart $\mathcal{M}'$ by simply replace their original token embeddings (\emph{e.g.}, word embeddings) with our length-aware tokens embeddings.

When training $\mathcal{M}'$, we can directly follow the training scheme of $\mathcal{M}$, like using the Teacher Forcing scheme in Eqn.~(\ref{eq:aic}) if $\mathcal{M}$ is autoregressive. During training, the length level embedding for level $l$ will only be trained with captions within a particular length range, thus the ``trait'' of image captions with different lengths is separately captured, enabling $\mathcal{M}'$ to perform length-aware modeling. During inference, apart from the image features, the length level embedding of the desired length level is also fed into $\mathcal{M}'$ as a control signal to generate image captions within the corresponding length range.

When setting the boundary \texttt{[$L_{low}$,$L_{high}$]} of a length level, we follow two simple principles: 1) there should be enough training data for each length level so that the length level embedding can be trained sufficiently; 2) the range of a length level should not be too narrow to ensure the flexibility of the generated captions. In our experiments, after checking the length distribution of captions in the MS COCO dataset, we explore two length level division plans which contain 4 and 5 length levels, respectively. As an example, the 4-level plan divides the image captions into 4 chunks with length inside the ranges \texttt{[1,9]}, \texttt{[10,14]}, \texttt{[15,19]}, and \texttt{[20,25]}, respectively, from rough to detailed. While the 5-level plan provides more fine-grained levels.

\subsubsection{Length-aware autoregressive caption decoders}

The decoder of most existing image captioning models can be broadly classified into two categories, \emph{i.e.}, LSTM-based and Transformer-based. 
Due to its simplicity, the proposed length level embedding can be easily integrated into these models.
Specifically, we implement it on AoANet~\cite{huang2019aoa} and VLP~\cite{zhou2019unified}, which are the most recent SOTA image captioning models of the two decoder categories, respectively.
Like many previous image captioning models, AoANet adopts an LSTM-based caption decoder. During each decoding step, the LSTM takes as input the concatenation of the word embedding of the input token and a context vector obtained from the image feature and the decoder context.
On the other hand, VLP uses a BERT~\cite{devlin2019bert}-style decoder that consists of a stack of Transformers. Specifically, VLP follows BERT and employs three types of embeddings to embed an input token, namely the word embedding, the positional embedding, as well as the segment type embedding.
To achieve length-controllable image captioning for these two methods, we directly add our length level embedding onto the word embedding of the input tokens for both AoANet and VLP, without any other modifications. Through this way, their caption decoders can explicitly model the length information of the input tokens.

\subsection{Non-autoregressive Length-controllable Decoding} \label{sec:naic}

\begin{figure}[t]
\centering
\includegraphics[width=1\textwidth]{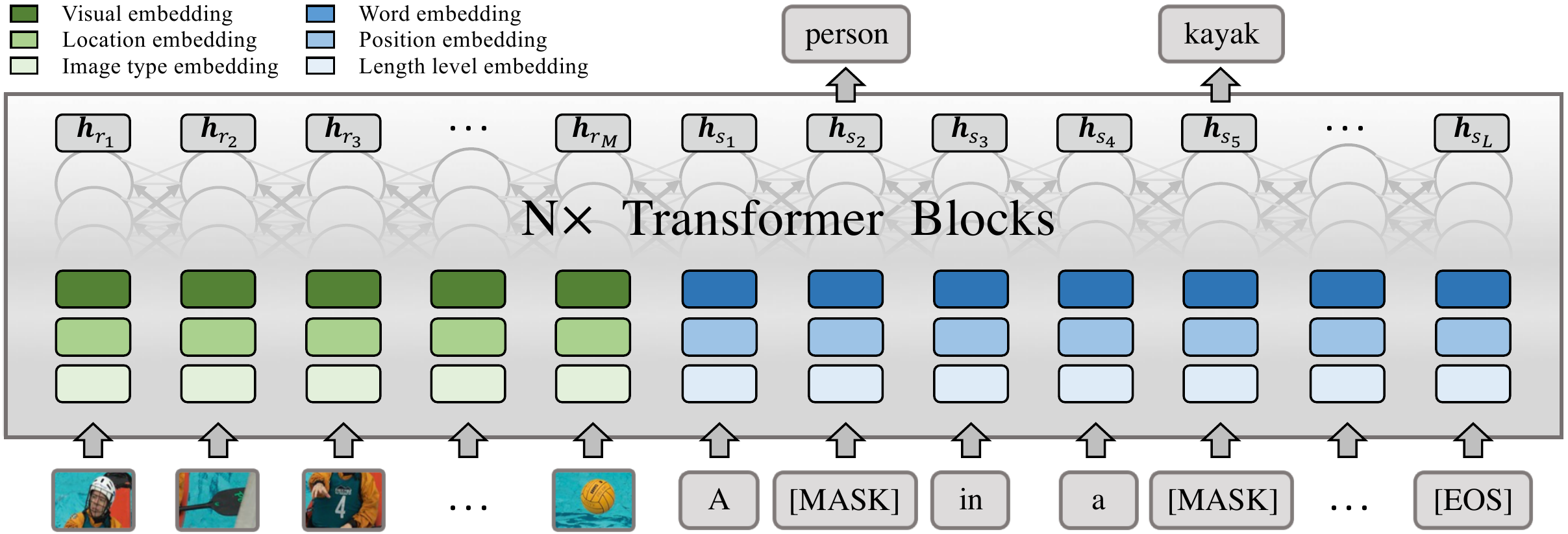}
\caption{The overview of LaBERT. Image regions and caption words are projected into the same dimensional space by the sum of three embeddings, respectively. All inputs are then combined together through $N\times$ Transformer blocks. The final hidden states $\bm{h}$ of \texttt{[MASK]} inputs are fed into a token classifier to predict their original tokens.}
\label{fig:arch}
\end{figure}

To improve the decoding efficiency for long image captions, we propose a non-autoregressive length-controllable image captioning model named as LaBERT, where we modify the embedding layer of BERT~\cite{devlin2019bert} to incorporate the image information and the length level information, as shown in Fig.~\ref{fig:arch}. Specifically, we follow~\cite{anderson2018bottom,zhou2019unified} and first adopt a pre-trained object detector to detect $M$ object proposals from $\bm{I}$, denoted as $\bm{R} = \{r_i\}_{i=1}^M$. The object detector is further employed to obtain the corresponding region features $\bm{F}_{e} = \{\bm{f}_{e,i}\}_{i=1}^M$, classification probabilities $\bm{F}_{c} = \{\bm{f}_{c,i}\}_{i=1}^M$, and localization features $\bm{F}_{l} = \{\bm{f}_{l,i}\}_{i=1}^M$ for $\bm{R}$. Similar to~\cite{zhou2019unified}, the input representation of $r_i$ is constructed by:
\begin{equation}\label{eq:xembed}
    \bm{x}_{r_i} = \bm{W}_e^T\bm{f}_{e,i} + \bm{W}_p^T[\text{LN}(\bm{f}_{c,i}),\text{LN}(\bm{f}_{l,i})] +
    \bm{e}_{img},
\end{equation}
where the first two term are the visual embedding and location embedding of $r_i$, respectively. $\bm{e}_{img} \in \mathbb{R}^{d}$ is a learnable embedding that differentiates the image regions from the text tokens. $[\cdot,\cdot]$ indicates the concatenate operation, and LN represents Layer Normalization \cite{ba2016layer}. $\bm{W}_e$ and $\bm{W}_p$ are two learnable projection matrices that project the corresponding features into $d$-D space.

\subsubsection{Training}
Given the target image caption $\bm{S}^*$, we first identify its length level $l$ and obtain the length range \texttt{[$L_{low}$,$L_{high}$]} of $l$. Then, we pad $\bm{S}^*$ with the \texttt{[EOS]} token to the longest length $L_{high}$. Following~\cite{ghazvininejad2019mask}, we construct the input sequence $\bm{S}$ by randomly replacing $m$ tokens in $\bm{S}^*$ with the \texttt{[MASK]} token, where $m$ is also randomly selected from the range \texttt{[1,$L_{high}$]}. 
Next, LaBERT attempts to predict the original tokens at all masked positions in $\bm{S}$ conditioned only on the embeddings of input image regions (obtained by Eqn.~(\ref{eq:xembed})) and the length-aware embeddings of the unmasked tokens in $\bm{S}$ (obtained by Eqn.~(\ref{eq:tembed})). Hence, the predicted conditional probabilities are independent with each other, allowing them to be calculated in parallel at inference time. We train LaBERT by minimizing the cross-entropy loss over all \emph{masked} positions:
\begin{equation} \label{eq:gLoss1}
\min \sum_{i=1}^{L_{high}}-\mathbbm{1}(s_i)\log p(s_i=s_i^*).
\end{equation}
$\mathbbm{1}(\cdot)$ is an indicator function whose value is $1$ if $s_i = \texttt{[MASK]}$ and $0$ otherwise. 

\subsubsection{Inference}

We perform parallel image caption decoding based on the idea of iterative refinement~\cite{ghazvininejad2019mask,lee2018deterministic}.
Specifically, at step $t=1$, we initialize the image caption $\bm{S}$ as $L_{high}$ consecutive \texttt{[MASK]} tokens. 
We first construct the input representations for text and image through Eqn.~(\ref{eq:tembed}) and Eqn.~(\ref{eq:xembed}), respectively. We then feed them into LaBERT to predict a probability distribution over a pre-defined vocabulary for every position in $\bm{S}$, denoted as $\bm{P}=\{p_i\}_{i=1}^{L_{high}}$. To encourage the model to predict longer captions, we propose to exponentially decay the probability of the \texttt{[EOS]} token by a factor $\gamma$ for predictions after $L_{low}$:
\begin{equation}\label{eq:eosdecay}
p_i(s_i=\texttt{[EOS]}) \gets \gamma^{L_{high}-i}p_i(s_i=\texttt{[EOS]}), ~ \forall i \in [L_{low}, L_{high}].
\end{equation}
Then, we obtain the refined $\bm{S}$ by updating all masked position:
\begin{equation} \label{eq:upds}
s_i\gets \arg\max_s p_i(s_{i}=s).
\end{equation}
Moreover, we obtain a confidence score $c_i = \max_s p_i(s_{i}=s)$ for each predicted token $s_i$, denoted as $\bm{C}=\{c_i\}_{i=1}^{L_{high}}$.

\begin{figure}[t]
\centering
\includegraphics[width=1.0\textwidth]{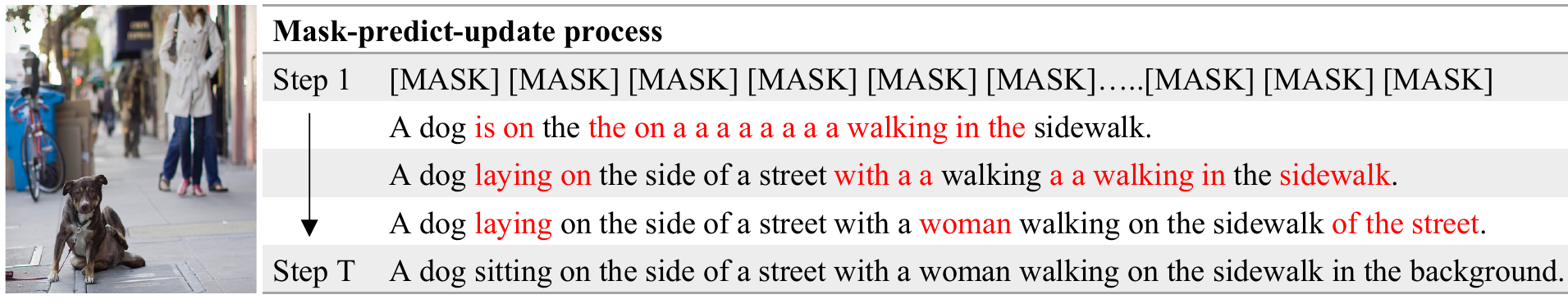}
\caption{An example from our experiments that illustrates the ``mask-predict-update'' process of LaBERT. At each step, all red tokens are masked and re-predicted in parallel, conditioned on other tokens in the sequence and visual information from the image.}
\label{fig:mask}
\end{figure}

At step $t=2$, we adopt the \textbf{mask}-\textbf{predict}-\textbf{update} procedure~\cite{ghazvininejad2019mask}, \emph{i.e.}, we find the lowest $n$ confidence scores in $\bm{C}$ and \textbf{mask} the corresponding positions in $\bm{S}$ with \texttt{[MASK]}. Next, the masked $\bm{S}$ will be fed into LaBERT to \textbf{predict} all masked positions in parallel, with the probability for \texttt{[EOS]} decayed by Eqn.~(\ref{eq:eosdecay}). We \textbf{update} all masked positions in $\bm{S}$ through Eqn.~(\ref{eq:upds}) to obtain the refined $\bm{S}$. And we propose to update $\bm{C}$ by:
\begin{equation} \label{eq:updc}
c_i \gets\left\{
\begin{aligned}
& \max_s p_i(s_i=s), & i~\text{is a masked position},\\
& (c_i + \max_s p_i(s_i=s)) / 2, & \text{otherwise}.
\end{aligned}
\right.
\end{equation}
The mask-predict-update procedure will be repeated until $t=T$ ($T$ can be smaller than $L_{high}$). The number of masks $n$ in each step is calculated by $n = \frac{T-t}{T}L_{high}$, which will decay linearly as the step $t$ increases. An illustration of the mask-predict-update procedure is shown in Fig.~\ref{fig:mask}.
\newline

Through iterative refinement, the computational complexity is decreased from ${\rm{\Theta}}(L_{high})$ in autoregressive methods to ${\rm{\Theta}}(T)$ in LaBERT. Moreover, the mistakes made at early steps in LaBERT are possible to be revised in the future steps, which is infeasible for autoregressive methods.
Note that the update rule in Eqn.~(\ref{eq:updc}) is different from the update rule in~\cite{ghazvininejad2019mask}, which only updates the confidence scores of the masked positions. In practice, we found ours (denoted as the \emph{global update rule}) performs much better in terms of caption quality. 
Besides, unlike many non-autoregressive text generation methods that rely on a length predictor to determine the length of the output at the start of decoding, our model is trained to automatically find a suitable end position within \texttt{[$L_{low}$,$L_{high}$]} for level $l$, thus this length predictor is not required. Moreover, LaBERT also allows dynamic length changes during the refinement process, while not using any additional insertion/deletion modules like in~\cite{gu2019levenshtein}.

\section{Experiments}

In this section, we first introduce the dataset and metrics we used in evaluation and the implementation details in Section~\ref{sec:dataset} and Section~\ref{sec:imp}, respectively. 
In the following sections, we verify the merit of the proposed length level embedding from two perspectives, \emph{i.e.}, the quality of generated image captions (Section~\ref{sec:ar} and~\ref{sec:nar}) and the controllability \& diversity (Section~\ref{sec:control}).
The verification is performed on two SOTA autoregressive models, and our non-autoregressive model LaBERT, to show the generalization ability of the length level embedding.
Meanwhile, we analyze the performance of LaBERT and discuss how it improves the efficiency of long captions generation in Section~\ref{sec:ana}.

\subsection{Dataset and Metrics} \label{sec:dataset}
We evaluate our proposed method on the popular MS COCO dataset~\cite{lin2014microsoft} which contains 123,287 images with at least 5 ground-truth captions for each image. We adopt the Karpathy's split setting~\cite{karpathy2015deep}, which uses 113,287 images for training, 5,000 for validation and 5,000 for offline evaluation. 

To evaluate the quality of the generated captions, we use standard metrics, including BLEU~\cite{papineni2002bleu}, ROUGE~\cite{Lin2004ROUGE}, METEOR~\cite{banerjee2005meteor}, CIDEr-D~\cite{vedantam2015cider}, and SPICE~\cite{anderson2016spice}. All these metrics except SPICE calculate the similarity between the reference and candidate image captions by considering their $n$-grams similarity. On the other hand, SPICE is based on scene-graph synonym matching which considers a scene-graph representation of an image by encoding objects, attributes, and relations. According to~\cite{anderson2016spice,kilickaya2017re}, SPICE and METEOR correlate best with human judgments in terms of caption quality among all these metrics. Moreover, since most ground-truth image captions in the \texttt{test} splits are short (more than 90\% contain 8-14 tokens), the performance of $n$-gram based metrics can be negatively affected when evaluating long candidate captions (\emph{e.g.}, longer than 14 tokens). Fortunately, SPICE is robust to the length of candidate captions, thus it should be the \emph{prior metric} for the evaluation of long captions.

\subsection{Implementation Details}\label{sec:imp}
For length-aware AoANet and VLP, we adopt their official codes as well as their experiment settings.
For LaBERT, we initialize it from the official pre-trained BERT-base~\cite{devlin2019bert} model, which have 12 layers of Transformer, 12 attention heads, and a hidden size of 768. 
We represent each input image as 100 object proposals extracted by a Faster RCNN~\cite{ren2015faster} pre-trained on the Visual Genome~\cite{krishna2017visual} dataset. We take the intermediate results at the \texttt{fc6} layer (2048-D) of the Faster RCNN as the region features $\bm{F}_{e}$. The classification labels $\bm{F}_{c}$ containing 1600 object categories are obtained from the final \texttt{softmax} layer.
The localization feature of each proposal is a 5-tuple contains the normalized coordinates of the top-left and bottom-right corner of the proposal and its relative area to the whole image.
We train LaBERT for 100,000 iterations with a batch size of 256. The AdamW~\cite{loshchilov2018fixing} optimizer is used with $\beta_1=0.9$, $\beta_2=0.999$, and a weight decay of 1e-2. We linearly warm-up the learning rate from 0 to 5e-5 over the first 1,000 iterations, and cosine decay it in the rest training steps. We use a label smoothing of 0.1, and a gradient clipping threshold of 1.0. The \texttt{[EOS]} decay factor $\gamma$ is determined by cross validation on the \texttt{val} splits for each level of LaBERT. Specifically, $\gamma$ is set to 0.88 and 0.95 for level 2 and level 3, respectively. For other length levels, we found LaBERT performs well without \texttt{[EOS]} decay.


\setlength{\tabcolsep}{7pt}
\begin{table}[t]
\centering
\caption{Performance of the length-aware version of AoANet and VLP on MS COCO Karpathy's \texttt{test} split. S, C, M and B@N, are short for SPICE, CIDEr-D, METEOR, and BLEU@N scores, respectively. The original results of AoANet and VLP are obtained from models trained by ourselves with the official codes and settings provided by the authors. All values are reported as a percentage ($\%$).}
\label{tab:ar}
\begin{tabular}{@{}c|l|cccc|cccc@{}}
\toprule
\rowcolor[HTML]{FFFFFF} 
\multicolumn{2}{l|}{Metrics}           &   S  &   C  &   M  &  B@4   &  S  &  C   &  M   &  B@4  \\ \midrule
\multicolumn{2}{l|}{Models}      &       \multicolumn{4}{c|}{AoANet} & \multicolumn{4}{c}{VLP} \\ \midrule
\multicolumn{2}{l|}{Original Results}  &  21.3   &  118.4   &  28.3   &  \textbf{36.9}   &  21.2   &  116.9   &   28.5  &  \textbf{36.5}  \\
\midrule
 \multirow{4}{*}{\rotatebox[origin=c]{90}{4-Level}} &Lv 1 (1-9)  &    19.6   &  107.4   &  25.9   &  33.1   &  18.9   &  103.0  &   25.2  &  31.8  \\
& Lv 2 (10-14)   &  21.7   &  117.6   &  28.6   &  35.8   &  21.4   &  \textbf{118.7}   &   28.8  &  36.0  \\
& Lv 3 (15-19)   &  22.7   &  79.9   &   28.7  &  26.6   &   22.4  &   92.5  &   \textbf{29.3}  &  28.4  \\
& Lv 4 (20-25)    &  22.7   &  29.5   &   27.7  &  20.2   &  22.4   &  40.0   & 28.5    &  21.9  \\ \midrule
 \multirow{5}{*}{\rotatebox[origin=c]{90}{5-Level}} &Lv 1 (1-9)      &  19.7   &  108.7   &  26.0   &  33.5   &  18.7   &  101.0   &  25.0   &  30.9  \\
& Lv 2 (10-13)    &  21.6   &  \textbf{118.8}   &  28.5   &  36.1   &  21.2   & 117.3    & 28.4    &  35.9  \\
& Lv 3 (14-17)    &  22.6   &   92.9  &   \textbf{29.0}  &   28.7  &  22.3   &  100.5   &  \textbf{29.3}   &   29.9 \\
& Lv 4 (18-21)    & \textbf{23.0}    &   48.4  &   28.2  &   22.7  &  22.4   &  60.4   &   28.7  &   24.0 \\
& Lv 5 (22-25)    & 22.9    &   18.9  &   27.2  &   18.8  &  \textbf{22.5}   &  28.1   &   28.1  &  20.3  \\ 
\bottomrule
\end{tabular}
\end{table}
\setlength{\tabcolsep}{1pt}

\subsection{Performance on AoANet and VLP}\label{sec:ar}

We first apply our length level embedding to two current SOTA models, \emph{i.e.}, AoANet~\cite{huang2019aoa} and VLP~\cite{zhou2019unified}. 
We train length-aware AoANet and VLP following their original training settings. During evaluation, we run our length-aware models on the \texttt{test} split multiple times, where each time we feed in a different length level embedding to generate the captions within different length ranges.
The results are recorded in Table~\ref{tab:ar}. From the table, on the normal length range (10-14 tokens), our 4-Level and 5-Level version of VLP and AoANet both achieve competitive or better performance than the original version. Our 4-level VLP even improve the CIDEr-D score by 1.8\% over the original VLP.
Note that, most of the image captions generated by original AoANet and VLP are inside the length range \texttt{[10,14]}, which indicates that our length-aware versions can maintain or even boost the performance of the original models on a normal length range.

The $n$-gram based metrics like CIDEr-D drops severely on longer length levels. However, as we discussed in Section~\ref{sec:dataset}, this does not mean the captions generated on these levels are bad. 
See Fig.~\ref{fig:example}, the 4-level VLP generates high-quality image captions on all levels. Specifically, on the shortest level, the image is concisely described, while on the longest level, 4-level VLP narrates the image in great detail, including visual concepts such as ``cheese'', ``spinach'' and ``tomatoes'', some are even missed in the ground-truth. Moreover, our models generally achieve remarkable SPICE scores for captions longer than 14 tokens. Our 5-level AoANet even achieves 23.0 SPICE score, which is 1.7\% higher than the original result. 
These results indicate that the length level embedding is well-suited for existing autoregressive image captioning models and makes them capable of producing high-quality results within different length ranges.

\setlength{\tabcolsep}{7pt}
\begin{table}[t]
\centering
\caption{Performance of LaBERT on MS COCO Karpathy's \texttt{test} split. R represents ROUGE. The results of AoANet and VLP are obtained from their papers.}
\label{tab:nar}
\begin{tabular}{@{}l|cccccccc@{}}
\toprule
{Metrics}    &   S  &   C  &   M  &  R   &  B@1   &  B@2   &  B@3   &  B@4  \\ 
 \midrule
 \multicolumn{9}{c}{State-Of-The-Art Models}\\
  \midrule
VLP~\cite{zhou2019unified}   &  21.2 & 116.9 & {28.4} & - & - & - & - & 36.5  \\
AoANet~\cite{huang2019aoa}   &  {21.3} & \textbf{119.8} & {28.4} & \textbf{57.5} & {77.4} & - & - & \textbf{37.2}  \\
\midrule
 \multicolumn{9}{c}{Single-Level LaBERT} \\
 \midrule
{Single Level (1-25)}      &  21.7   &  116.8   &  27.9   &  57.0   &  77.4   &  61.0   &   46.5  &  35.0  \\
  \midrule
 \multicolumn{9}{c}{4-Level LaBERT} \\
 \midrule
{lvl 1 (1-9)}      &  19.5   &  101.6   &  25.4   &  54.7   &  72.5   &  56.3   &   41.8  &  30.0  \\
{lvl 2 (10-14)}    &  21.8   &  118.2   &  28.4   &  57.4   &  \textbf{77.6}  &  \textbf{61.3}   &   \textbf{46.9}  &  35.3  \\
{lvl 3 (15-19)}    &  \textbf{22.3}   &  90.5   &   \textbf{28.6}  &  53.1   &   66.8  &  50.6  &   37.0  &  26.8  \\
{lvl 4 (20-25)}    &  22.2   &  39.9   &   27.7  &  46.9   &  56.1   &  40.9   & 28.6    &  19.9  \\
\bottomrule
\end{tabular}
\end{table}
\setlength{\tabcolsep}{1pt}

\subsection{Performance on LaBERT} \label{sec:nar}

We further apply the length level embedding to our proposed non-autoregressive image captioning model, LaBERT. To make comparisons, we implement a single-level version of LaBERT whose length range is \texttt{[1,25]}. Moreover, the number of refine steps for 4-level LaBERT is set to 10, 15, 20 and 25, for level 1-4, respectively, so that we can compare LaBERT with the autoregressive image captioning models under roughly the same decoding complexity. Analogously, The total decoding steps of the single-level LaBERT is set to 25.
The results are shown in Table~\ref{tab:nar}. Compared with single-level LaBERT, 4-level LaBERT achieves clearly better performance on all metrics on the second level, which coincides with our experiments in Section~\ref{sec:ar}. Moreover, 4-level LaBERT yields significantly higher SPICE scores on level 3 and level 4. These results demonstrate the advantage of the length level embedding in our non-autoregressive image caption model.
Moreover, under the same computational complexity, 4-level LaBERT performs comparably with 4-level AoANet and VLP as well as the SOTA results, which demonstrates the effectiveness of length-level embedding in non-autoregressive image caption decoding.

\subsection{Performance Analysis of LaBERT} \label{sec:ana}

\setlength{\tabcolsep}{7.5pt}
\begin{table}[t]
\caption{Performance analysis on 4-level LaBERT. ``Speedup'' indicates the relative speedup over an autoregressive baseline.}
\label{tab:abla}
\begin{center}
\begin{tabular}{@{}c|l|ccccccc@{}}
\toprule
\multicolumn{2}{l|}{Method} & S & C & M & R & B@1 & B@4 & Speedup\\
 \midrule
 \multirow{6}{*}{\rotatebox[origin=c]{90}{Lvl 4}}&10 refine steps  & 21.6 & 39.5 & 27.3 & 46.9 & 55.9 & 19.0 & $\times 2.5$\\
&12 refine steps & 21.9 & 39.8 & 27.5 & 46.7 & 56.1 & 19.4 & $\times 2.1$\\
&15 refine steps & 22.0 & 39.5 & 27.5 & 46.6 & 55.9 & 19.3 & $\times 1.67$\\
&20 refine steps & 22.4 & 39.9 & 27.8 & 47.0 & 56.1 & 19.8 & $\times 1.25$\\
&25 refine steps & 22.2 & 39.9 & 27.7 & 46.9 & 56.1 & 19.9 & $\times 1.0$\\
& 4-Level VLP    & 22.4 & 40.0 & 28.5 & 47.0 & 56.0 & 21.9 & $\times 1.0$\\
 \midrule
\multirow{2}{*}{\rotatebox[origin=c]{90}{Lvl 2}}&w/o global update & 21.7 & 116.6 & 28.2 & 57.2 & 77.3 & 34.8 & - \\
&w/o \texttt{[EOS]} decay & 21.3 & 116.0 & 27.8 & 57.1 & 78.3 & 34.9 & - \\ 
&Original Results & 21.8 & 118.2 & 28.4 & 57.4 & 77.6 & 35.3 & - \\
\bottomrule
\end{tabular}
\end{center}
\end{table}
\setlength{\tabcolsep}{1pt}

In this section, we analyze the effect of some hyper-parameters in 4-level LaBERT, \emph{i.e.}, the number of refine steps $T$, the \texttt{[EOS]} decay factor $\gamma$ in Eqn.~(\ref{eq:eosdecay}), and the global update rule in Eqn.~(\ref{eq:updc}). To analyze the effect of refine steps $T$, we vary $T$ from 10 to 25 for caption generation on the fourth level; to show the importance of the \texttt{[EOS]} decay factor $\gamma$ and the global update rule, we perform ablation experiments for them on the second level. The results are shown in Table~\ref{tab:abla}. From the table, the autoregressive baseline, \emph{i.e.}, 4-level VLP performs slightly better on SPICE; however, it requires 25 steps to decode the entire sequence. On the contrary, our non-autoregressive LaBERT can use a much smaller $T$ to achieve acceleration: by reducing $T$ from 25 to 15, we speedup the decoding by $1.67\times$ with only a minor performance degradation on SPICE (0.2\%). We can acquire further speedup ($2.1\times$) by setting $T=12$, with a small sacrifice on SPICE (0.3\%). Nevertheless, the performance obtain by LaBERT with 10-20 refine steps are still competitive to the 25-step performance as well as the performance of the 4-level VLP, which verifies the capability of LaBERT in efficient image captioning decoding.
Moreover, the global update rule and the \texttt{[EOS]} decay are shown to be important for LaBERT. After removing them, the CIDEr-D score of LaBERT on the second level drops by 1.6\% and 2.2\%, respectively.

\subsection{Controllability and Diversity Analysis}\label{sec:control}

\begin{figure}[t]
\centering
\includegraphics[width=1.0\textwidth]{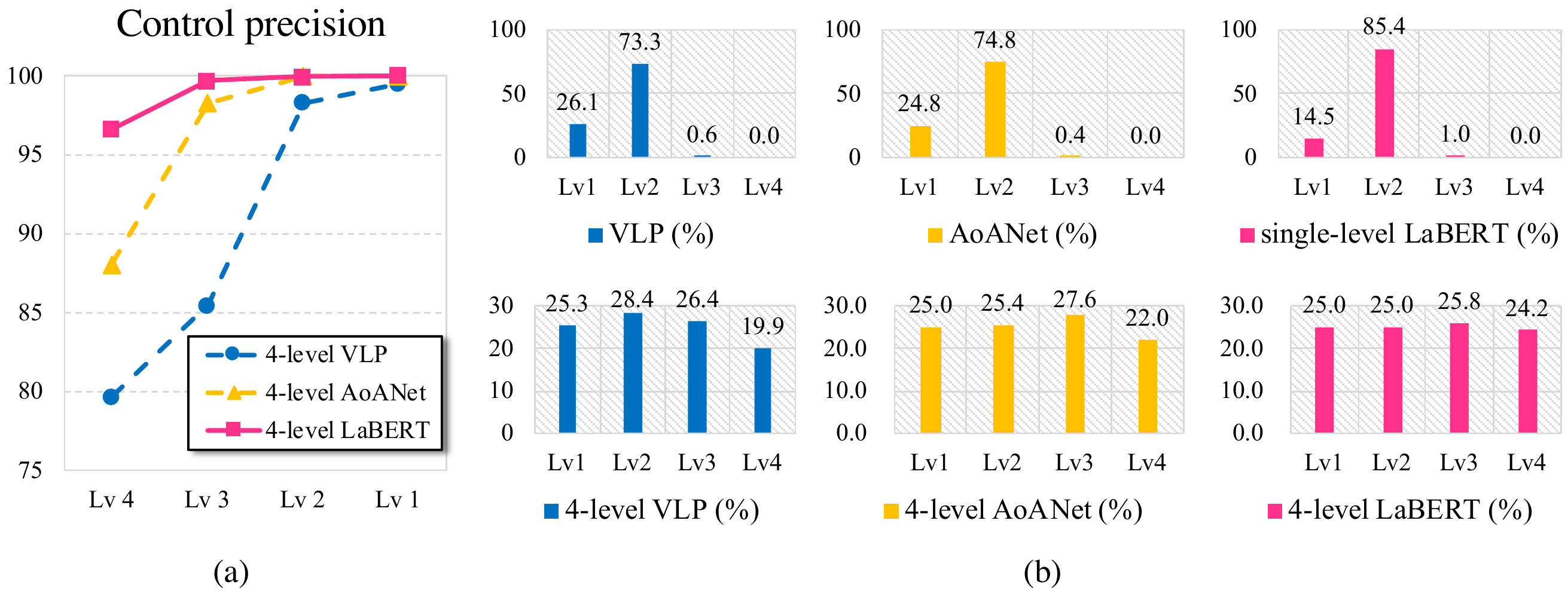}
\caption{Analysis of controllability and diversity on \texttt{test} split. (a) The control precision of our 4-level version of AoANet, VLP and LaBERT; (b) The length distributions of image captions generated by our 4-level length-aware models and their counterparts.}
\label{fig:length}
\end{figure}

In this section, we further analyze the ``control precision'' of the length level embedding, \emph{i.e.}, given a length level embedding, the probability of generating image captions within the desired length range. 
We calculate the control precision for the 4-level version of AoANet, VLP and LaBERT, and present the results in Fig.~\ref{fig:length}(a). As shown in the figure, all methods accurately control the length of the generated image captions, and our non-autoregressive model, LaBERT, yields the best control precision (more than 95\%) among all levels. This result verifies the effectiveness of the proposed length level embedding in generating length-controllable image captions. Besides, the control precision drops on longer levels, which may due to the lack of long captions in the MS COCO dataset.

We also perform diversity analysis for the image captions generated by different models, as shown in Fig.~\ref{fig:length}(b) and Table~\ref{tab:div}. From Fig.~\ref{fig:length}(b), the length of the image captions generated by our length-aware models are uniformly distributed among all length levels. On the contrary, the results of the original AoANet, VLP and the single-level LaBERT distribute mainly in the shortest two levels. We further evaluate the diversity of the image captions on $n$-gram diversity metrics like Div-1 and Div-2, as well as the recently proposed SelfCIDEr~\cite{wang2019describing} score that focuses on semantic diversity.
From Table~\ref{tab:div}, our 4-level models perform clearly better on all metrics, which means we can obtain diverse captions for an image with our length-aware image captioning models. Interestingly, our non-autoregressive model LaBERT significantly outperforms all compared autoregressive methods on all three diversity metrics.

\setlength{\tabcolsep}{8pt}
\begin{table}[t]
\caption{Diversity analysis. BS denotes beam search with a beam size of 4.}
\label{tab:div}
\centering
\begin{tabular}{@{}lccccc@{}}
\toprule
\multirow{2}{*}{Models} & \multicolumn{2}{c}{AoANet} & \multicolumn{2}{c}{VLP} & {LaBERT} \\ \cmidrule(l){2-6} 
             & BS & 4-Level & BS & 4-Level & 4-Level  \\ \midrule
SelfCIDEr~\cite{wang2019describing}    & 0.590  &  0.689  &  0.623  & 0.762  & \textbf{0.841}   \\
Div-1    & 0.291  &  0.378  &  0.313  & 0.406  & \textbf{0.411}   \\
Div-2    & 0.462  &  0.523  &  0.470  & 0.559  & \textbf{0.575}   \\
\bottomrule
\end{tabular}
\end{table}
\setlength{\tabcolsep}{1pt}

\section{Conclusion}

In this paper, we propose to use a length level embedding for length-controllable image captioning. By simply adding our length level embedding on the word embeddings of input tokens, we endow existing image captioning methods with the ability to control the length of their predictions. To improve the decoding efficiency of long captions, we further propose a non-autoregressive image captioning model, LaBERT, that generates image captions in a length-irrelevant complexity. The experiments demonstrate the effectiveness of the proposed method. 
\let\thefootnote\relax\footnotetext{\textbf{Acknowledgments} This work was partially supported by the Key-Area Research and Development Program of Guangdong Province (2018B010107001), National Natural Science Foundation of China 61836003 (key project), Guangdong Project 2017ZT07X183, Fundamental Research Funds for the Central Universities D2191240.}

%
%
\bibliographystyle{splncs04}
\bibliography{eccv2020submission}

\begin{thebibliography}{10}
\providecommand{\url}[1]{\texttt{#1}}
\providecommand{\urlprefix}{URL }
\providecommand{\doi}[1]{https://doi.org/#1}

\bibitem{anderson2016spice}
Anderson, P., Fernando, B., Johnson, M., Gould, S.: Spice: Semantic
  propositional image caption evaluation. In: European Conference on Computer
  Vision. pp. 382--398. Springer (2016)

\bibitem{anderson2018bottom}
Anderson, P., He, X., Buehler, C., Teney, D., Johnson, M., Gould, S., Zhang,
  L.: Bottom-up and top-down attention for image captioning and visual question
  answering. In: Proceedings of the IEEE Conference on Computer Vision and
  Pattern Recognition. pp. 6077--6086 (2018)

\bibitem{ba2016layer}
Ba, J.L., Kiros, J.R., Hinton, G.E.: Layer normalization. stat  \textbf{1050},
  ~21 (2016)

\bibitem{banerjee2005meteor}
Banerjee, S., Lavie, A.: Meteor: An automatic metric for mt evaluation with
  improved correlation with human judgments. In: Proceedings of the acl
  workshop on intrinsic and extrinsic evaluation measures for machine
  translation and/or summarization. pp. 65--72 (2005)

\bibitem{bengio2015scheduled}
Bengio, S., Vinyals, O., Jaitly, N., Shazeer, N.: Scheduled sampling for
  sequence prediction with recurrent neural networks. In: Advances in Neural
  Information Processing Systems. pp. 1171--1179 (2015)

\bibitem{chen2020say}
Chen, S., Jin, Q., Wang, P., Wu, Q.: Say as you wish: Fine-grained control of
  image caption generation with abstract scene graphs. In: Proceedings of the
  IEEE/CVF Conference on Computer Vision and Pattern Recognition. pp.
  9962--9971 (2020)

\bibitem{chen2018factual}
Chen, T., Zhang, Z., You, Q., Fang, C., Wang, Z., Jin, H., Luo, J.:
  ``factual''or``emotional'': Stylized image captioning with adaptive learning
  and attention. In: Proceedings of the European Conference on Computer Vision
  (ECCV). pp. 519--535 (2018)

\bibitem{chen2015microsoft}
Chen, X., Fang, H., Lin, T.Y., Vedantam, R., Gupta, S., Doll{\'a}r, P.,
  Zitnick, C.L.: Microsoft coco captions: Data collection and evaluation
  server. arXiv preprint arXiv:1504.00325  (2015)

\bibitem{cornia2019show}
Cornia, M., Baraldi, L., Cucchiara, R.: Show, control and tell: a framework for
  generating controllable and grounded captions. In: Proceedings of the IEEE
  Conference on Computer Vision and Pattern Recognition. pp. 8307--8316 (2019)

\bibitem{dai2017towards}
Dai, B., Fidler, S., Urtasun, R., Lin, D.: Towards diverse and natural image
  descriptions via a conditional gan. In: Proceedings of the IEEE International
  Conference on Computer Vision. pp. 2970--2979 (2017)

\bibitem{deshpande2019fast}
Deshpande, A., Aneja, J., Wang, L., Schwing, A.G., Forsyth, D.: Fast, diverse
  and accurate image captioning guided by part-of-speech. In: Proceedings of
  the IEEE Conference on Computer Vision and Pattern Recognition. pp.
  10695--10704 (2019)

\bibitem{devlin2019bert}
Devlin, J., Chang, M.W., Lee, K., Toutanova, K.: Bert: Pre-training of deep
  bidirectional transformers for language understanding. In: Proceedings of the
  2019 Conference of the North American Chapter of the Association for
  Computational Linguistics: Human Language Technologies, Volume 1 (Long and
  Short Papers). pp. 4171--4186 (2019)

\bibitem{gan2017stylenet}
Gan, C., Gan, Z., He, X., Gao, J., Deng, L.: Stylenet: Generating attractive
  visual captions with styles. In: Proceedings of the IEEE Conference on
  Computer Vision and Pattern Recognition. pp. 3137--3146 (2017)

\bibitem{gehring2017convolutional}
Gehring, J., Auli, M., Grangier, D., Yarats, D., Dauphin, Y.N.: Convolutional
  sequence to sequence learning. In: International Conference on Machine
  Learning. pp. 1243--1252 (2017)

\bibitem{ghazvininejad2019mask}
Ghazvininejad, M., Levy, O., Liu, Y., Zettlemoyer, L.: Mask-predict: Parallel
  decoding of conditional masked language models. In: Proceedings of the 2019
  Conference on Empirical Methods in Natural Language Processing and the 9th
  International Joint Conference on Natural Language Processing (EMNLP-IJCNLP).
  pp. 6114--6123 (2019)

\bibitem{goodfellow2014generative}
Goodfellow, I., Pouget-Abadie, J., Mirza, M., Xu, B., Warde-Farley, D., Ozair,
  S., Courville, A., Bengio, Y.: Generative adversarial nets. In: Advances in
  neural information processing systems. pp. 2672--2680 (2014)

\bibitem{Greff2015LSTM}
Greff, K., Srivastava, R.K., Koutník, J., Steunebrink, B.R., Schmidhuber, J.:
  Lstm: A search space odyssey. IEEE Transactions on Neural Networks \&
  Learning Systems  \textbf{28}(10),  2222--2232 (2015)

\bibitem{gu2017non}
Gu, J., Bradbury, J., Xiong, C., Li, V.O., Socher, R.: Non-autoregressive
  neural machine translation. arXiv preprint arXiv:1711.02281  (2017)

\bibitem{gu2019levenshtein}
Gu, J., Wang, C., Zhao, J.: Levenshtein transformer. In: Advances in Neural
  Information Processing Systems. pp. 11179--11189 (2019)

\bibitem{huang2019aoa}
Huang, L., Wang, W., Chen, J., Wei, X.Y.: Attention on attention for image
  captioning. In: The IEEE International Conference on Computer Vision (ICCV)
  (October 2019)

\bibitem{karpathy2015deep}
Karpathy, A., Fei-Fei, L.: Deep visual-semantic alignments for generating image
  descriptions. In: Proceedings of the IEEE conference on computer vision and
  pattern recognition. pp. 3128--3137 (2015)

\bibitem{kikuchi2016controlling}
Kikuchi, Y., Neubig, G., Sasano, R., Takamura, H., Okumura, M.: Controlling
  output length in neural encoder-decoders. In: Proceedings of the 2016
  Conference on Empirical Methods in Natural Language Processing. pp.
  1328--1338 (2016)

\bibitem{kilickaya2017re}
Kilickaya, M., Erdem, A., Ikizler-Cinbis, N., Erdem, E.: Re-evaluating
  automatic metrics for image captioning. In: Proceedings of the 15th
  Conference of the European Chapter of the Association for Computational
  Linguistics: Volume 1, Long Papers. pp. 199--209 (2017)

\bibitem{krishna2017visual}
Krishna, R., Zhu, Y., Groth, O., Johnson, J., Hata, K., Kravitz, J., Chen, S.,
  Kalantidis, Y., Li, L.J., Shamma, D.A., et~al.: Visual genome: Connecting
  language and vision using crowdsourced dense image annotations. International
  Journal of Computer Vision  \textbf{123}(1),  32--73 (2017)

\bibitem{lee2018deterministic}
Lee, J., Mansimov, E., Cho, K.: Deterministic non-autoregressive neural
  sequence modeling by iterative refinement. In: Proceedings of the 2018
  Conference on Empirical Methods in Natural Language Processing. pp.
  1173--1182 (2018)

\bibitem{li2019entangled}
Li, G., Zhu, L., Liu, P., Yang, Y.: Entangled transformer for image captioning.
  In: Proceedings of the IEEE International Conference on Computer Vision. pp.
  8928--8937 (2019)

\bibitem{Lin2004ROUGE}
Lin, C.Y.: Rouge: A package for automatic evaluation of summaries. In: In
  Proceedings of the Workshop on Text Summarization Branches Out (WAS 2004)
  (2004)

\bibitem{lin2014microsoft}
Lin, T.Y., Maire, M., Belongie, S., Hays, J., Perona, P., Ramanan, D.,
  Doll{\'a}r, P., Zitnick, C.L.: Microsoft coco: Common objects in context. In:
  European conference on computer vision. pp. 740--755. Springer (2014)

\bibitem{liu2018controlling}
Liu, Y., Luo, Z., Zhu, K.: Controlling length in abstractive summarization
  using a convolutional neural network. In: Proceedings of the 2018 Conference
  on Empirical Methods in Natural Language Processing. pp. 4110--4119 (2018)

\bibitem{loshchilov2018fixing}
Loshchilov, I., Hutter, F.: Fixing weight decay regularization in adam  (2018)

\bibitem{mathews2018semstyle}
Mathews, A., Xie, L., He, X.: Semstyle: Learning to generate stylised image
  captions using unaligned text. In: Proceedings of the IEEE Conference on
  Computer Vision and Pattern Recognition. pp. 8591--8600 (2018)

\bibitem{mathews2016senticap}
Mathews, A.P., Xie, L., He, X.: Senticap: Generating image descriptions with
  sentiments. In: Thirtieth AAAI conference on artificial intelligence (2016)

\bibitem{papineni2002bleu}
Papineni, K., Roukos, S., Ward, T., Zhu, W.J.: Bleu: a method for automatic
  evaluation of machine translation. In: Proceedings of the 40th annual meeting
  on association for computational linguistics. pp. 311--318. Association for
  Computational Linguistics (2002)

\bibitem{ren2015faster}
Ren, S., He, K., Girshick, R., Sun, J.: Faster r-cnn: Towards real-time object
  detection with region proposal networks. In: Advances in neural information
  processing systems. pp. 91--99 (2015)

\bibitem{rennie2017self}
Rennie, S.J., Marcheret, E., Mroueh, Y., Ross, J., Goel, V.: Self-critical
  sequence training for image captioning. In: Proceedings of the IEEE
  Conference on Computer Vision and Pattern Recognition. pp. 7008--7024 (2017)

\bibitem{shetty2017speaking}
Shetty, R., Rohrbach, M., Anne~Hendricks, L., Fritz, M., Schiele, B.: Speaking
  the same language: Matching machine to human captions by adversarial
  training. In: Proceedings of the IEEE International Conference on Computer
  Vision. pp. 4135--4144 (2017)

\bibitem{shuster2019engaging}
Shuster, K., Humeau, S., Hu, H., Bordes, A., Weston, J.: Engaging image
  captioning via personality. In: Proceedings of the IEEE Conference on
  Computer Vision and Pattern Recognition. pp. 12516--12526 (2019)

\bibitem{stern2019insertion}
Stern, M., Chan, W., Kiros, J., Uszkoreit, J.: Insertion transformer: Flexible
  sequence generation via insertion operations. In: International Conference on
  Machine Learning. pp. 5976--5985 (2019)

\bibitem{sutskever2014sequence}
Sutskever, I., Vinyals, O., Le, Q.V.: Sequence to sequence learning with neural
  networks. In: Advances in neural information processing systems. pp.
  3104--3112 (2014)

\bibitem{sutton2000policy}
Sutton, R.S., McAllester, D.A., Singh, S.P., Mansour, Y.: Policy gradient
  methods for reinforcement learning with function approximation. In: Advances
  in neural information processing systems. pp. 1057--1063 (2000)

\bibitem{vaswani2017attention}
Vaswani, A., Shazeer, N., Parmar, N., Uszkoreit, J., Jones, L., Gomez, A.N.,
  Kaiser, {\L}., Polosukhin, I.: Attention is all you need. In: Advances in
  neural information processing systems. pp. 5998--6008 (2017)

\bibitem{vedantam2015cider}
Vedantam, R., Lawrence~Zitnick, C., Parikh, D.: Cider: Consensus-based image
  description evaluation. In: Proceedings of the IEEE conference on computer
  vision and pattern recognition. pp. 4566--4575 (2015)

\bibitem{vinyals2015show}
Vinyals, O., Toshev, A., Bengio, S., Erhan, D.: Show and tell: A neural image
  caption generator. In: Proceedings of the IEEE conference on computer vision
  and pattern recognition. pp. 3156--3164 (2015)

\bibitem{wang2018semi}
Wang, C., Zhang, J., Chen, H.: Semi-autoregressive neural machine translation.
  In: Proceedings of the 2018 Conference on Empirical Methods in Natural
  Language Processing. pp. 479--488 (2018)

\bibitem{wang2017diverse}
Wang, L., Schwing, A., Lazebnik, S.: Diverse and accurate image description
  using a variational auto-encoder with an additive gaussian encoding space.
  In: Advances in Neural Information Processing Systems. pp. 5756--5766 (2017)

\bibitem{wang2019describing}
Wang, Q., Chan, A.B.: Describing like humans: on diversity in image captioning.
  In: Proceedings of the IEEE Conference on Computer Vision and Pattern
  Recognition. pp. 4195--4203 (2019)

\bibitem{wu2017image}
Wu, Q., Shen, C., Wang, P., Dick, A., van~den Hengel, A.: Image captioning and
  visual question answering based on attributes and external knowledge. IEEE
  transactions on pattern analysis and machine intelligence  \textbf{40}(6),
  1367--1381 (2017)

\bibitem{xu2015show}
Xu, K., Ba, J., Kiros, R., Cho, K., Courville, A., Salakhudinov, R., Zemel, R.,
  Bengio, Y.: Show, attend and tell: Neural image caption generation with
  visual attention. In: International conference on machine learning. pp.
  2048--2057 (2015)

\bibitem{yang2019auto}
Yang, X., Tang, K., Zhang, H., Cai, J.: Auto-encoding scene graphs for image
  captioning. In: Proceedings of the IEEE Conference on Computer Vision and
  Pattern Recognition. pp. 10685--10694 (2019)

\bibitem{yao2018exploring}
Yao, T., Pan, Y., Li, Y., Mei, T.: Exploring visual relationship for image
  captioning. In: Proceedings of the European conference on computer vision
  (ECCV). pp. 684--699 (2018)

\bibitem{yao2019hierarchy}
Yao, T., Pan, Y., Li, Y., Mei, T.: Hierarchy parsing for image captioning. In:
  Proceedings of the IEEE International Conference on Computer Vision. pp.
  2621--2629 (2019)

\bibitem{yao2017boosting}
Yao, T., Pan, Y., Li, Y., Qiu, Z., Mei, T.: Boosting image captioning with
  attributes. In: Proceedings of the IEEE International Conference on Computer
  Vision. pp. 4894--4902 (2017)

\bibitem{zhou2019unified}
Zhou, L., Palangi, H., Zhang, L., Hu, H., Corso, J.J., Gao, J.: Unified
  vision-language pre-training for image captioning and vqa. arXiv preprint
  arXiv:1909.11059  (2019)

\end{thebibliography}
\end{document}